%% file: main.tex
\documentclass[sigconf,natbib=false,nonacm]{acmart}

\input{header}
\AtBeginDocument{%
  }

\setcopyright{acmlicensed}
\copyrightyear{2025}
\acmYear{2025}
\acmDOI{XXXXXXX.XXXXXXX}

\acmConference[EASE 2026]{The 30th International Conference on Evaluation and Assessment in Software Engineering}{9–12 June, 2026}{Glasgow, Scotland, United Kingdom}

\RequirePackage[
  datamodel=acmdatamodel,
  style=acmnumeric,
  ]{biblatex}

\begin{document}

\title{FLOPs vs Real Work: The Importance of Replication in AI Efficiency Assessment}

\author{Enrique Barba Roque}
\email{enrique@ebarba.com}
\orcid{0000-0002-7018-498X}
\affiliation{%
  \institution{Delft University of Technology}
  \city{Delft}
  \country{The Netherlands}
}

\author{Luís Cruz}
\orcid{0000-0002-1615-355X}
\email{l.cruz@tudelft.nl}
\affiliation{%
  \institution{Delft University of Technology}
  \city{Delft}
  \country{The Netherlands}
}

\renewcommand{\shortauthors}{Barba Roque et al.}

\begin{abstract}
AI efficiency has recently taken the spotlight in both academy and industry due to massive model scales, high energy demands, and environmental costs. While reporting Floating Point Operations (FLOPs) is a traditional approach for assessing computational costs, the relationship between FLOPs and execution time is not straightforward, as layers with the same number of FLOPs may not have the same execution time because some operations are more easily parallelized than others. This paper sets out to replicate the original experiments from a study that proposed the $\alpha-FLOPs$ estimation formula to verify whether the results remain applicable on newer, more powerful hardware.

During the replication process, we identify limitations in the replication materials provided by the original study, including a lack of specific dependency details and transparency regarding regression data. Our results validate the thesis that raw FLOPs alone are not an appropriate metric for execution time, as spatial dimensions remain more easily parallelized than kernel dimensions. However, fine-grained measurements reveal that the relationship is much less straightforward than previously shown, with newer hardware exhibiting instabilities and discontinuities in execution time, including jumps and oscillations, that the $\alpha-FLOPs$ formula generally underestimates. Ultimately, this work validates the empirical findings from the original study but shows negative results when applying the $\alpha-FLOPs$ estimation. We also highlight the critical need for complete and accurate replication packages for research on hardware-dependent efficiency assessment and provide a complete replication package for our implementation to facilitate further study.
\end{abstract}



\received{30 April 2026}

\maketitle

\section{Introduction}
AI efficiency has recently taken the spotlight in both academy and industry~\cite{10.1145/3639475.3640097,10.1145/3381831,tan2019efficientnet}.
AI capabilities come from massive scales: for example, Large Language Models (LLMs) have billions of parameters trained on terabytes of open-source code from platforms such as GitHub~\cite{lozhkov2024starcoder}.
This scale brings significant computational challenges, including the need for dedicated hardware for parallel computation, such as GPUs \cite{DBLP:conf/mlsys/WuRGAAMCBHBGGOM22}, along with high energy demands and environmental costs~\cite{10.1145/3381831}.
The International Energy Agency projects that AI energy consumption will double over the next 5 years \cite{IEA2025EnergyAI}, further straining existing electrical infrastructure and increasing greenhouse gas emissions.
Given this scale, it is highly relevant to put focus on AI optimization to reduce inference time and energy consumption.

%

To optimize AI models, developers need accurate metrics to assess relative costs and model efficiency. 
While execution time or energy consumption can be accurate metrics, they can be difficult to measure during model development, since they require specialized software profilers and access to bare metal \cite{10.1145/3743095.3743099,DBLP:journals/corr/abs-2506-09599}.
The traditional alternative to reporting the computational approach of an AI model is to report Floating Point Operations (FLOPs)~\cite{flopsToEnergy}.
This metric is easy to derive from the model's architecture by counting the number of matrix multiplications and their dimensions, and gives a sense of computational costs: models with higher FLOPs need to perform more multiplications, which directly affects time and energy consumption.


However, the relationship between FLOPs and execution time is not as straightforward. The work by \citeauthor{flopsCriticism}~\cite{flopsCriticism} examines this relationship for Convolutional Neural Networks.
Their work runs multiple experiments for different convolutional configurations that introduce FLOPs across different dimensions, such as input size, 
input and output channels or kernel size.
Their results show that layers with the same number of FLOPs do not necessarily have the same execution time, because some convolutional operations and dimensions are parallelized more easily than others.
To mitigate this, the paper proposes an estimation formula, $\alpha-FLOPs$, derived from their empirical measurements.
The $\alpha-FLOPs$ formula can provide reasonable execution-time estimates by running preliminary measurements and performing a regression.

The findings of their manuscript have valuable implications for Software Engineering for AI (SE4AI).
Providing developers with accurate, hardware-aware execution-time estimates supports informed decision-making during model design.
The proposed approach has the potential to be adapted and extended for different AI architectures to identify similar parallelization patterns in other AI models, and the $\alpha-FLOPs$ formula could be extended to account for these different architectures.
However, one of the limitations intrinsic to this approach is that it is highly hardware-dependent.
The method requires preliminary measurements for each specific GPU, and the estimation formula was derived from empirical measurements in the hardware available to the authors.
Given the rapid evolution of AI-specific hardware, some of the original findings may no longer be applicable due to new hardware optimizations or architectures.

Therefore, we set out to replicate the original experiments from the study to verify whether the results remain applicable.
During the replication process, we identified limitations in the replication materials provided by the study.
The provided replication package contains only the data and code to plot it, but no code to run experiments.
Additionally, there are other limitations, such as the absence of specific dependency details (e.g., AI libraries and GPU drivers), a lack of transparency about the data used for regression, and inconsistencies between the paper details and practical implementations.

Concretely, we define the following replication objectives:
\begin{itemize}
    \item \textbf{RO1.} Replicate the finding that raw FLOPs are not a sufficient proxy for CNN execution time, even when configurations have equal theoretical FLOPs.
    \item \textbf{RO2.} Replicate the finding that execution time depends on where FLOPs are introduced, with FLOPs in spatial dimensions being more parallelizable than FLOPs in kernel dimensions.
    \item \textbf{RO3.} Replicate the finding that the $\alpha$-FLOPs correction can provide accurate execution-time estimates after hardware-specific regression.
\end{itemize}

The contributions of this paper can be summarized as follows: (1) replicate the experiments from the original study \cite{flopsCriticism} in newer hardware and identify how efficiency patterns change with hardware evolution; (2) replicate the $\alpha-FLOPs$ estimation method and evaluate its validity for our hardware; (3) analyze the impact of poor replication packages when attempting to reproduce hardware-dependent results; and (4) a complete replication package of the original experiments, available in Zenodo\footnote{\url{https://doi.org/10.5281/zenodo.18842300}}.

\section{Background}
The intuition behind using FLOPs for comparing costs of neural network models is simple: the model needs to perform a certain number of operations for a forward pass, and each of these operations takes some amount of time $t_{FLOP}$.
\begin{equation}
    E_{Total} = t_{FLOP} \cdot FLOPs
\end{equation}

However, this approach is naive. Mainly, it does not account for the effect of parallelization. When running multiple operations in parallel, their costs are not additive. Instead, depending on the nature of the operations and the capacity of the GPU, parallel multiplications will take as much time and energy as a single multiplication. According to Gustafson's law \cite{10.1145/42411.42415}, for a task with a sequential and a parallel component, the sequential component takes the same time, while the parallel component's cost decreases with hardware capacity.
Therefore, different neural networks with the same theoretical FLOP count can have different energy consumption.

The work by \citeauthor{flopsCriticism} shows this in practice for shows. In their study, the authors measure the execution times of convolutional layers across different hyperparameter combinations.
For CNNs, these hyperparameters are the input dimensions, the number of channels, and the kernel size. 
In their experiments, they measure forward time for different hyperparameter combinations, increasing some parameters while reducing others, to control and maintain the same number of FLOPs.

Results show that layers with the same theoretical FLOPs exhibit vastly different execution times, depending on the hyperparameters and the dimension at which those FLOPs are applied.
They observe that FLOPs are easily parallelizable for spatial dimensions. They find a significant reduction in execution time as the input size increases and the channel size decreases.
On the other hand, increasing the kernel dimension while reducing the size or the number of channels leads to higher execution time for layers with the same FLOPs.

Based on their results, the study proposes a correction for FLOPs, where the estimated execution time for a layer is expressed as:
\begin{equation}
    T = \alpha_k(S) \cdot FLOPS
\end{equation}
\begin{equation} \label{eq:alpha-flops}
    \alpha_k(S) = \left(\frac{S_k + \beta_k(S-S_k)}{S}\right)^{\gamma_k}
\end{equation}

In this formula, $S_k$ is defined as ``a parameter that grows slowly with K, with $S_1 = 1$''.
Parameters $\beta_k$ and $\gamma_k$ are determined via regression using empirical data from different convolutional configurations.
These parameters are hardware-dependent, but the paper notes that, based on a preliminary study, $\beta_k$ and $\gamma_k$ show little variance across different hardware.
This formula provides reasonable estimates of CNN execution time for the study's setup.
The paper demonstrates the validity of the estimation by plotting the FLOPs-based estimate alongside the empirical results, showing that the estimate lies close to the original values.

The contributions of the original paper can be summarized as follows: a) the paper empirically demonstrates that FLOPs alone are not a good measure of CNN execution time. b) the relationship between execution time and FLOPs depends on the specific hyperparameter configuration, and how parallelizable those FLOPs are. c) provides an empirically obtained formula that can estimate the execution time of a convolutional layer based on FLOPs and hyperparameters.
In addition to the paper and its contributions, the authors provide a dataset with the data for the measurements presented in the manuscript\footnote{\url{https://github.com/asperti/alpha_flops_dataset}}.
These contributions paint a clear picture of FLOPs' limited validity as a metric for assessing the computational cost of an AI model.
The correction formula provided can also be adapted and applied to different AI architectures, which motivates our attempt to replicate the results.

\section{Limitations to Replicability}
The findings of the original study are valuable for the field of efficiency and energy assessment of AI and CNNs.
A similar empirical methodology could be extended and applied to different AI architectures to identify FLOP and execution-time patterns across their hyperparameters.
However, since the approach is empirical and hardware-dependant, we need to verify that it remains applicable to a newer GPU, motivating this replication.

During this replication, we identified several limitations in the replication materials provided in the original paper. 
The dataset provided partially covers the necessary information, but it is far from a replication package, as it contains only the data needed to recreate the plots.
The authors mention that they `` do not provide the obvious code for timing execution or for regression over data ''.
Additionally, the paper does not include other relevant details for replication, such as the library or driver versions used in the experiment, code or details for crucial steps in this methodology, the data used for regression, and vague definitions of certain parameters in the formula, among others.

This section presents in detail the limitations encountered when attempting to replicate the results of this paper on newer hardware and how these limitations hinder replicability.
These limitations will inform the methodology for our own replication, which will include several assumptions to address the missing information.

\paragraph{Lack of dependency specification} 
Neither the paper nor the dataset specifies which of the two major AI libraries (TensorFlow or PyTorch) is used to implement the convolutional layers, or the versions used.
Additionally, the authors do not mention GPU drivers or CUDA versions used.
All of these details introduce variability in the methodology. Newer AI libraries might introduce optimizations on how layers and matrices are processed, reducing execution time for the same hardware.

\paragraph{Missing dataset and configuration for regression} 
Before being able to compute time estimations, one needs to define the hardware-specific parameters $\beta_k$ and $\gamma_k$ from the $\alpha-FLOPs$ function in \autoref{eq:alpha-flops}. To do this, the original study uses nonlinear regression on data collected from a sample of convolutional configurations.
The authors define the data as triplets $(S, C, K)$, along with the execution time $T$, where $S$ is the input size, $C$ is the channel size, and $K$ is the kernel size.
In this regard, the paper lacks information about the sampling strategy used or the range of values tested for these parameters.
This definition also does not clarify the range for the 2D values. For a 2D convolution, size S is defined by width and height, $S = W \times H$; the number of input and output channels can also be different, $C = C_{in} \times C_{out}$, and the kernel size is also 2-dimensional, $K = K_1 \times K_2$.
Matrix multiplication efficiency in t a GPU depends not only on the total number of elements but also on the dimensionality~\cite{10.5555/3455716.3455964}.
Additionally, the paper does not provide information on the configuration used for the non-linear regression, such as the library or functions used, the initial values for $\beta_k$ and $\gamma_k$, and the lower and upper bounds for these parameters.

\paragraph{Inconsistencies between paper and code}
The dataset provided with the paper includes a function to compute execution-time predictions using the $\alpha-FLOPs$ formula, with the values reported in the paper. However, this computation contains two inconsistencies with respect to the formulas stated in the manuscript.
First, the number of FLOPs for an input of width and height $W,H$, and input channels $C_{in}$, and a convolutional layer with kernel size $K_1 \times K_2$ and output channels $C_{out}$ is computed as $FLOPs = \mathbf{2} \cdot K_1 \cdot K_2 \cdot C_{in} \cdot W \cdot H \cdot C_{out}$.
However, the function provided in the artifact does not use the constant term $2$.
While this is just a scaling factor, it can lead to inconsistencies if it is used in the regression but not in the estimations.
Second, after the computation $T = \alpha_k(S) \cdot FLOPS$, the code multiplies this value by a scaling factor called \emph{final}, $T = final \cdot T$.
This value is set to $final = 0.0375$ in the artifact.
There is no justification for it. We conjecture that this is an additional regression parameter or a manually tuned value to adjust the prediction to the real measurements as closely as possible.

\input{methodology}

\section{Results}

\subsection{Regression Results}
After running the regression on our collected data, we obtain the following values for $\beta_K$, $\gamma_K$, and $final_K$. The values obtained for our hardware are $\beta_1 = 0.00972514$, $\gamma_1 = 1.1779239$ and $final_1 = 0.00549217$ for $K = 1$, and $\beta_K = 0.00785036$, $\gamma_K = 1.0601711$ and $final_K = 0.00401956$ for $K > 1$.

\subsection{Experiments Results}
In the following sections, we present the results for the replications of the original experiments. We visualize results as plots, with average times on the y-axis and the independent variable on the x-axis.
We also show the Root Mean Square Error (RMSE) and the normalized RMSE by range for each configuration.
The datapoints do not include error bars, since the Inter Quartile Range of all time measurements is negligible, at least 3 orders of magnitude smaller than actual measurements.

\subsubsection{Constant FLOPs}

\begin{figure}[t]
    \centering
    \includegraphics[width=\linewidth]{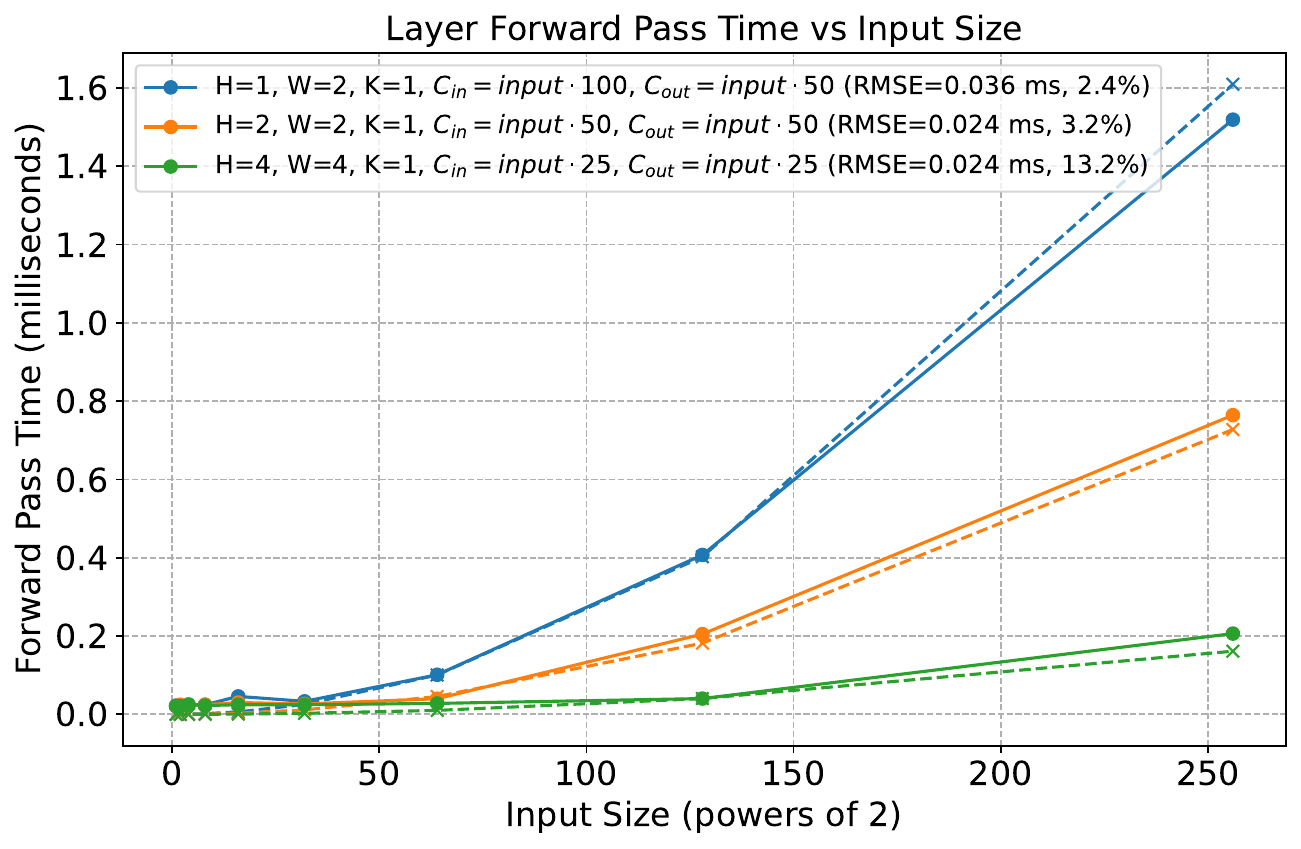}
    \caption{Experiment A: Execution time and estimations for increasing channels with different spatial dimensions. Estimations are represented with the dashed line.}
    \label{fig:exp-a}
\end{figure}

\autoref{fig:exp-a} shows the forward pass execution time for increasing input and output channels, for configurations from Experiment A. Each colored line represents a different set of configurations, and dashed lines show the time estimation using the $\alpha-FLOPs$ formula.
Each of the points in the same vertical line has the same number of theoretical FLOPs: the FLOPs introduced by a higher input size are compensated for by using a lower number of input and output channels.
If FLOPs alone were a good proxy for execution time, all of the points in the same vertical line should take the same amount of time to complete a forward pass.
Instead, the figure shows that FLOPs are more computationally intensive when introduced in the channel dimension, whereas they can be more easily parallelized across spatial dimensions.
The estimation provided by the $\alpha-FLOPs$ formula lies close to the actual measurements, within $\sim 5 \%$ difference, overestimating for smaller input sizes and underestimating for larger inputs.
RMSE is between $0.02$ and $0.04$ ms, or between $2.4\% - 13\%$ normalized for the measurement range for the three configurations
which is acceptable given that measurements lie between $0$ to $1.5$ ms.

These results are consistent with those of the original study.
For the original experiments, execution times range from $\sim 0.2$ to $8$ milliseconds due to significantly less powerful hardware, while our results show a $\sim 10\times$ speedup.
However, the evolution of execution time with increasing size follows the same pattern and aligns well with the original estimation function.
In this scenario, our setup behaves as observed in the original study.

\begin{figure}[t]
    \centering
    \includegraphics[width=\linewidth]{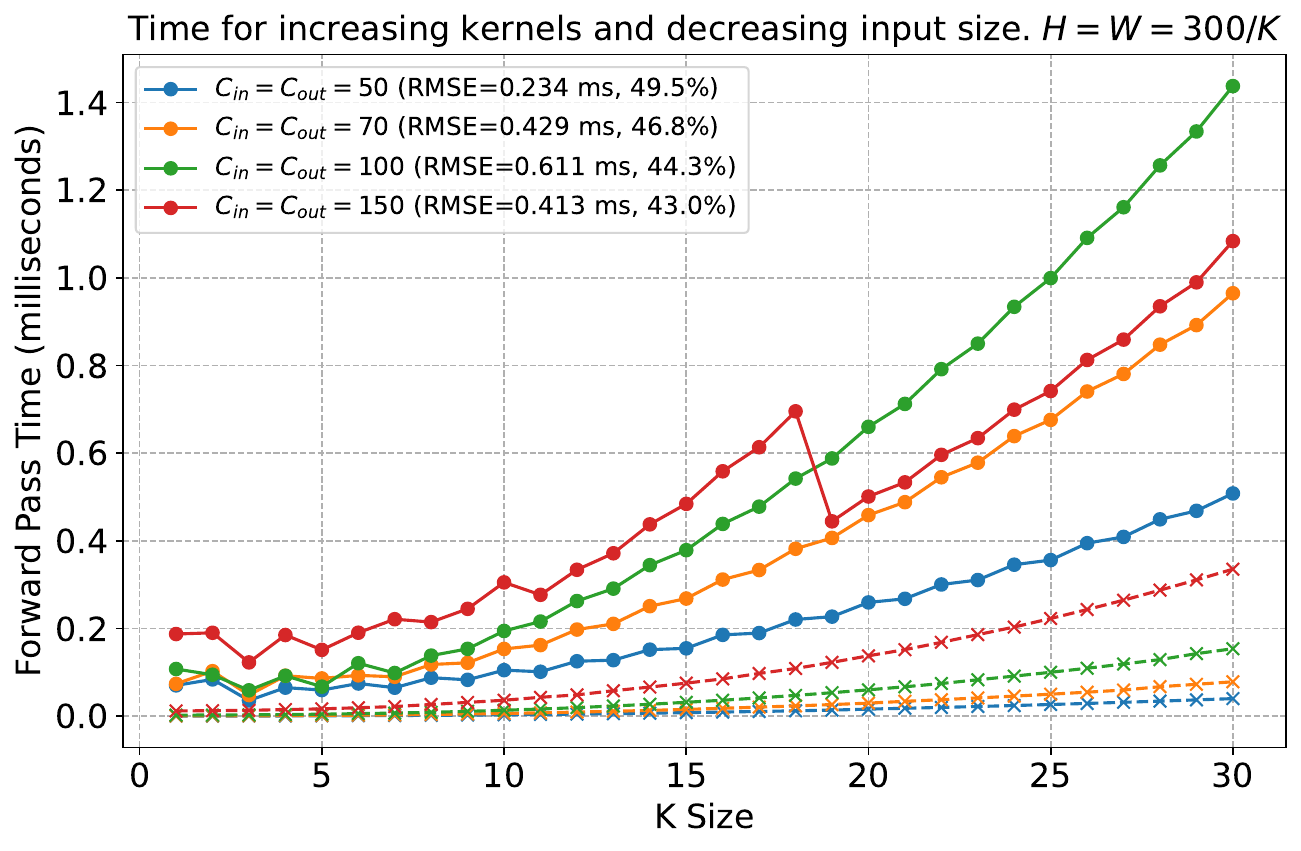}
    \caption{Experiment B: Execution time and estimations for increasing kernels and decreasing spatial dimensions. Estimations are represented with the dashed line.}
    \label{fig:exp-b}
\end{figure}

\autoref{fig:exp-b} plots the results for Experiment B. 
In this figure, the input size is fixed, and the kernel size $K$ (x-axis) increases while the input size dimensions decrease at a fixed ratio, maintaining $H=W=300/K$. 
Each configuration set, represented by a color, uses a different number of input and output channels. Thanks to the ratio rule, each point of a configuration set has the same theoretical FLOPs. 
If FLOPs alone provided a good estimate of execution time, each colored line would be approximately horizontal, with FLOPs increasing only as the number of channels increases.
Instead, the execution time increases consistently as $K$ increases and the number of dimensions decreases, following an approximately quadratic trend. This again supports the original findings that convolutional operations are more readily parallelizable across spatial dimensions than across kernel dimensions. 
One exception to this pattern occurs for $C_{in} = C_{out} = 150$, which exhibits a notable reduction in execution time for values $K>19$, with $H=W=\operatorname{round}(300/19) = 16$.
At this point, the execution time decreases by approximately $47\%$, from $0.69\,\text{ms}$ to $0.44\,\text{ms}$, after which the trend continues from this lower baseline, remaining below the curve for $C_{in} = C_{out} = 100$, despite the latter being theoretically less computationally demanding.
The estimates provided by the $\alpha$-FLOPs formula reproduce the qualitative shape of the empirical data but fail to match its scale, remaining distant from the measured results.
RMSE is between $0.2$ and $0.6$ ms, corresponding to around $47\%$ deviation from the actual measurement range.

Compared to the results from the original study, our empirical measurements follow similar trends, albeit with the same $\sim 10\times$ speedup.
However, we did not achieve the same level of accuracy in our hardware for the $\alpha-FLOPs$ estimations.
Additionally, the decrease observed for the $C_{in} = C_{out} = 150$ does not happen in the original study.

\begin{figure}[t]
    \centering
    \includegraphics[width=\linewidth]{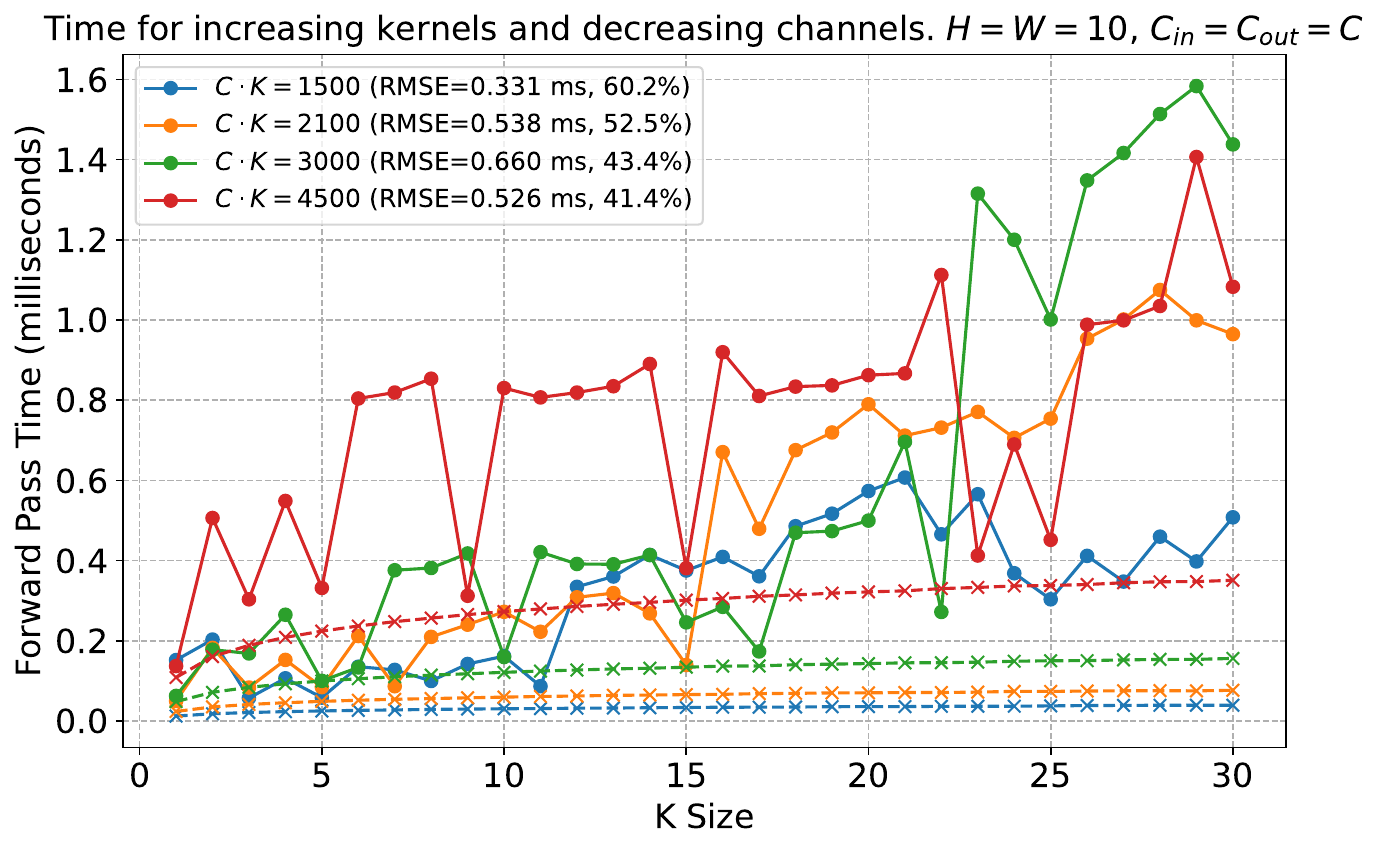}
    \caption{Experiment C: Execution time and estimations for increasing kernels and decreasing channels. Estimations are represented with the dashed line.}
    \label{fig:exp-c}
\end{figure}

\autoref{fig:exp-c} shows the results for Experiment C. Similarly to the previous experiment, the kernel size (x-axis), but now the input size is fixed, while the number of input and output channels decreases at a fixed ratio $R$, maintaining $C \cdot K = R$. 
This way, each set of configurations, represented by a color, has the same theoretical FLOPs, which increase only as the ratio increases.
Sets with higher ratios have higher theoretical FLOPs, and their execution times should consistently exceed those of smaller configurations.
However, the results for these experiments are inconsistent. While the most expensive set ($C \cdot K = 4500$) is the most expensive for lower values of $K$, it gets overtaken by smaller sets for larger values of $K$.
The estimation for these runs consistently underestimates execution time and does not adapt well to measurement evolution.

The results of this experiment differ from those of the original study.
The behavior of these convolutional configurations observed in the previous study is much more stable.
The execution time for increasing kernel sizes and decreasing channels shows approximately constant behavior, close to the theoretical behavior, and the differences between sets with different $R$ values are consistent, with lower $R$ values remaining below larger $R$ values.

\subsubsection{Increasing FLOPs Experiments}



\begin{figure*}[t]
    \centering
    \begin{subfigure}[b]{0.48\textwidth}
        \centering
        \includegraphics[width=\linewidth]{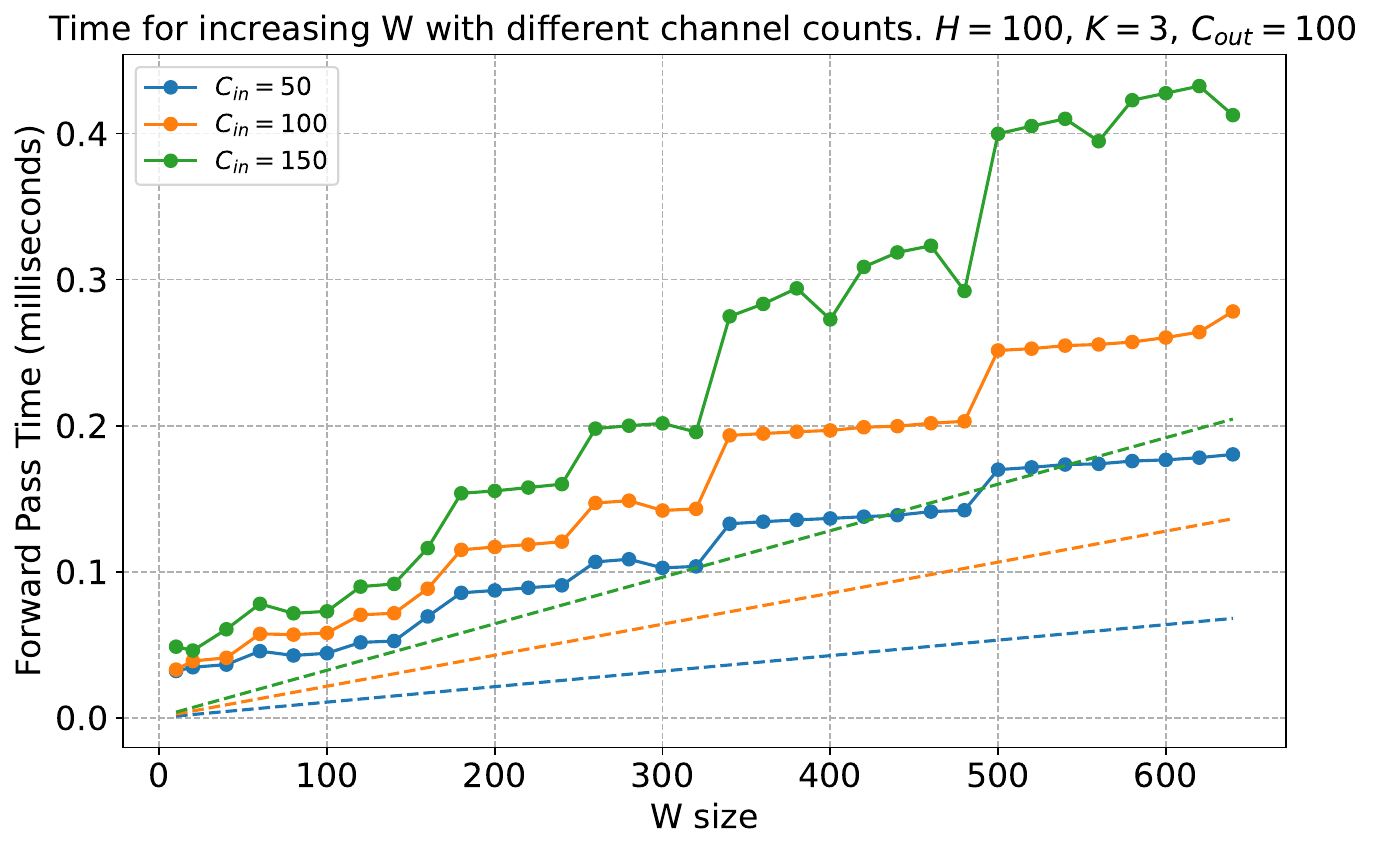}
        \caption{Limited sampling: 33 datapoints}
        \label{fig:exp-D}
    \end{subfigure}
    \hfill
    \begin{subfigure}[b]{0.48\textwidth}
        \centering
        \includegraphics[width=\linewidth]{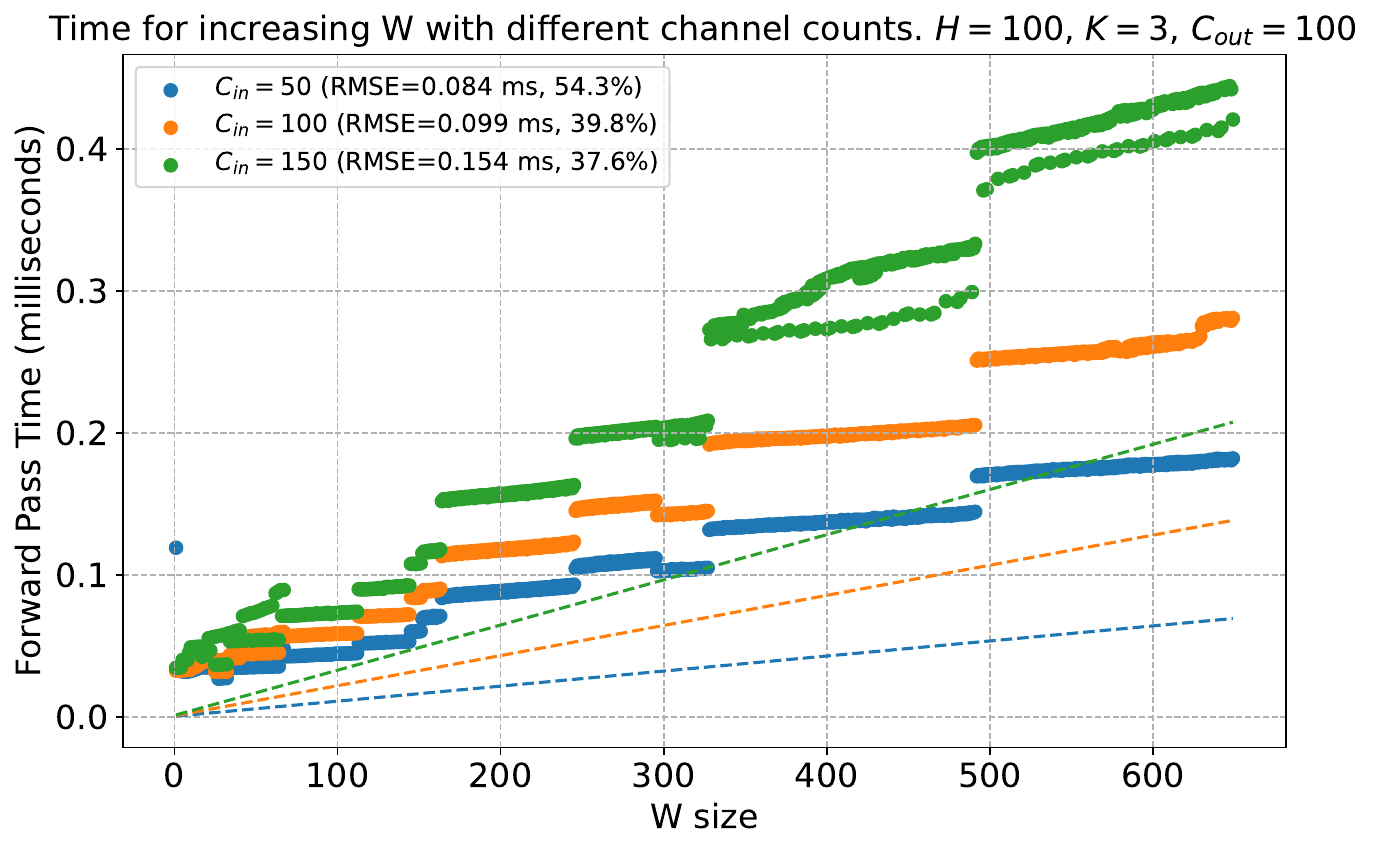}
        \caption{Full sampling: 649 datapoints}
        \label{fig:exp-DE}
    \end{subfigure}
    \caption{Experiment D: Execution time and estimations for increasing input size and different input channels. Estimations are represented with the dashed line.}
    \label{fig:exp-D-combined}
\end{figure*}

\autoref{fig:exp-D-combined} presents the results for Experiment D, where one input dimension $W$ acts as the independent variable, for three different values for input channels, and with the rest of the hyperparameters fixed.
\autoref{fig:exp-D} shows the results using the original resolution, with $W$ increased in steps of 20
Considering only theoretical FLOPs as a proxy, execution time should increase linearly with FLOPs.
Instead, in these configurations, execution time seems to increase in discrete steps for specific values of $W$, with a small linear increase between steps.
The estimate from our $\alpha-FLOPs$ implementation underestimates the actual measurements by about $50\%$, though it follows the same increasing trend. RMSE is up to $1.5$ ms for the worst case, and predictions deviate between $39\% - 56\%$ according to normalized RMSE.

\autoref{fig:exp-DE} shows the same setup with a higher resolution of samples, measuring for every $W$ value between 1 and 650.
This figure shows that the increase in execution time is not as stable as it might seem when sampling only a few values.
This resolution helps identify where the step jumps occur, with the largest ones occurring for values $W = \{164, 246, 328, 492\}$.
The figure also reveals oscillations for large values of $C_{in} = 150$, where convolutional configurations for certain values of $W$ perform faster than others.

Empirical measurements from the previous study for this experiment are closer to the theoretical model. 
In the original experiment, execution time increases linearly with input size $W$.
The two larger configurations show a similar step-up for $W = 420$.
In comparison, our results are more erratic, both at the lower resolution used in the original paper and at our extended resolution.

\begin{figure*}[t]
    \centering
    \begin{subfigure}[b]{0.48\textwidth}
        \centering
        \includegraphics[width=\linewidth]{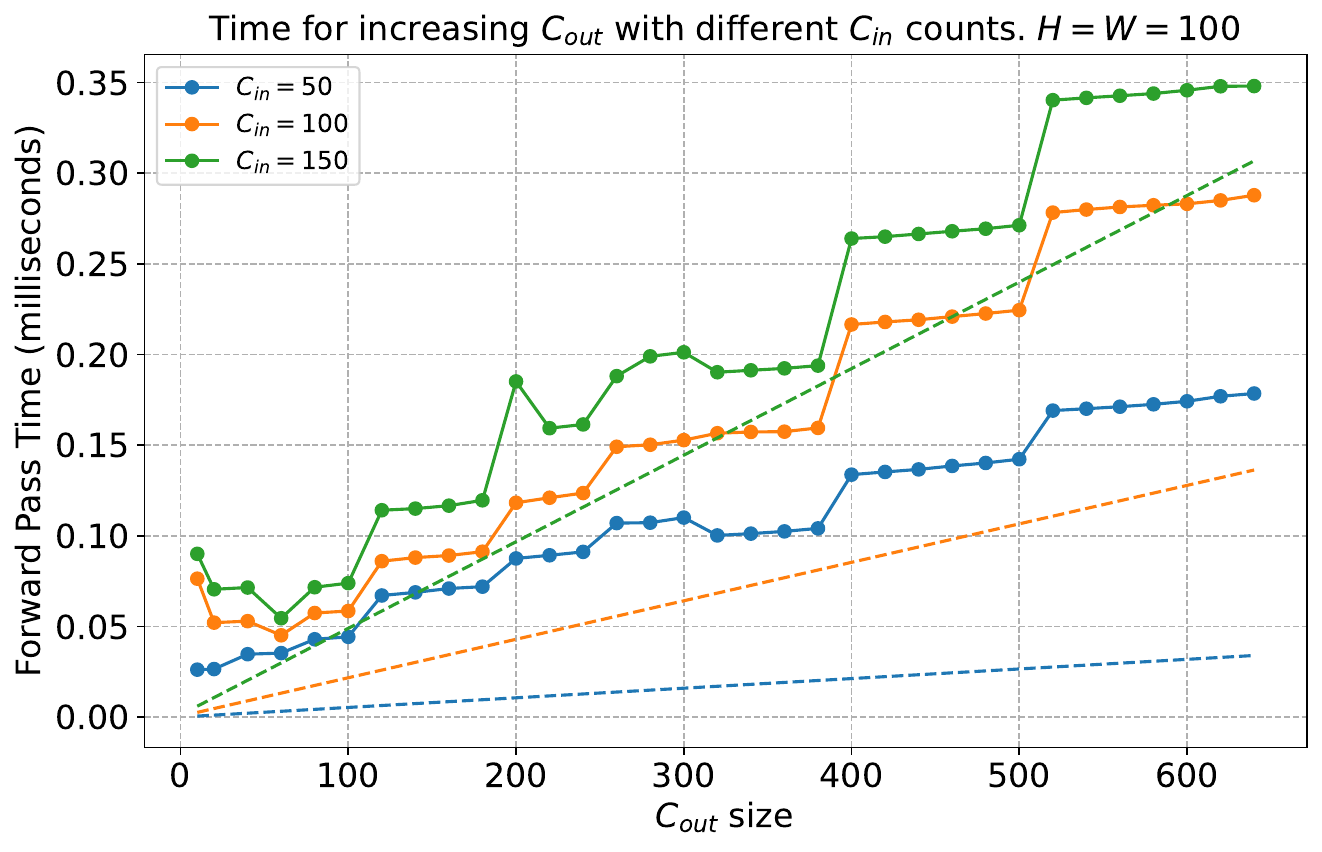}
        \caption{Limited sampling: 33 datapoints}
        \label{fig:exp-E}
    \end{subfigure}
    \hfill
    \begin{subfigure}[b]{0.48\textwidth}
        \centering
        \includegraphics[width=\linewidth]{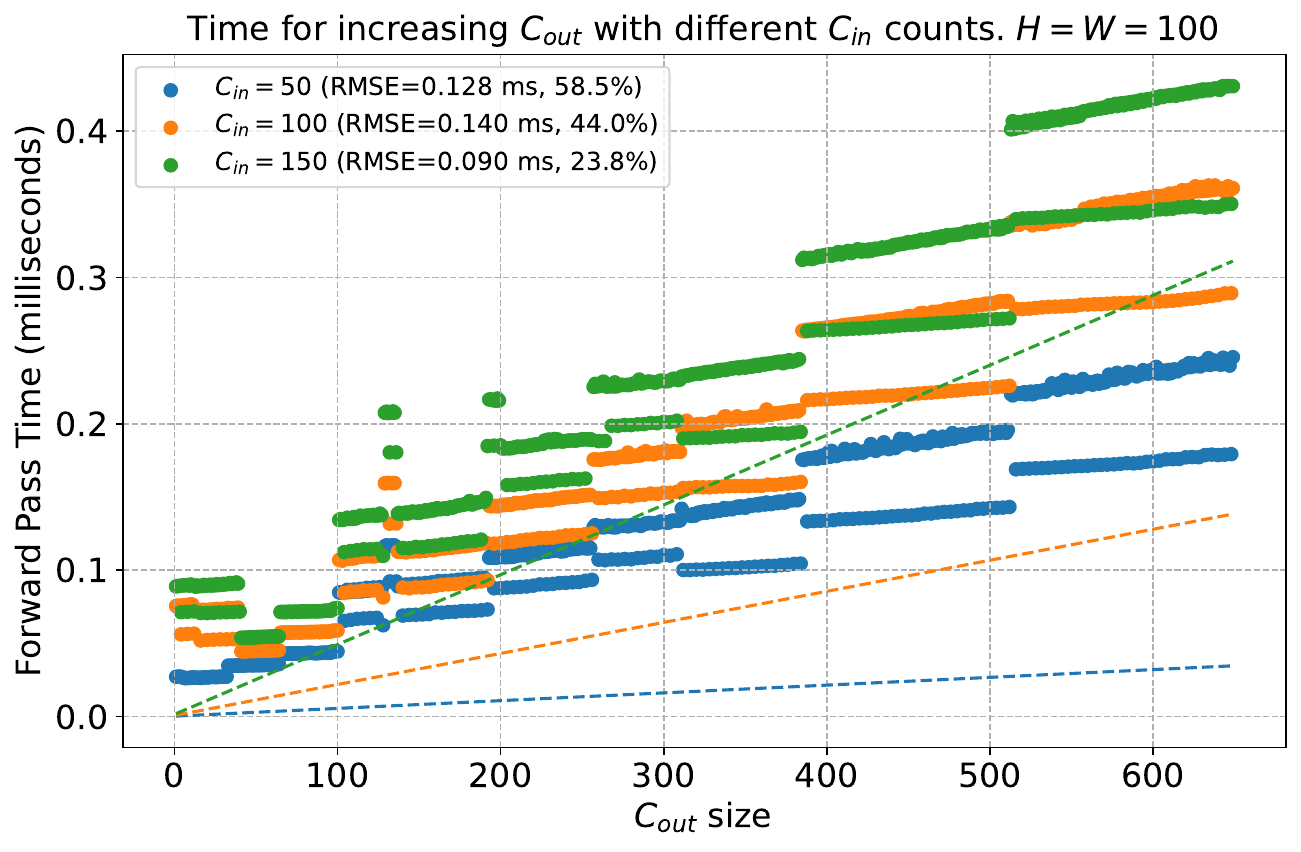}
        \caption{Full sampling: 649 datapoints}
        \label{fig:exp-EE}
    \end{subfigure}
    \caption{Experiment E: Execution time and estimations for increasing output channels and different input channels. Estimations are represented with the dashed line.}
    \label{fig:exp-E-combined}
\end{figure*}

\autoref{fig:exp-E-combined} shows the results of Experiment E. For this experiment, the independent variable is the number of output channels $C_{out}$, for three different values of input channels $C_{in}$.
\autoref{fig:exp-E} shows a lower resolution with 33 datapoints.
The behavior is similar to the previous figure: execution time increases stepwise for certain values of $C_{out}$, with a marginal increase between steps.
The $\alpha-FLOPs$ formula once again underestimates real execution time, showing larger errors for smaller layers but better predictions for larger configurations, with normalized RMSE ranging from $19\%$ to $64\%$.

\autoref{fig:exp-EE} shows the maximum resolution with all possible values of $C_{out}$. The configurations in this experiment are much more unstable than in the previous experiment and reveal behaviors that are hidden at lower resolution.
In this case, the values of $C_{out}$ that trigger the larger step ups are $C_{out} = \{101, 192, 257, 385, 513\}$.
Additionally, two bands of values are distinguishable for the three values of $C_{in}$ at every step.
Zooming in on the figure shows that the lower band is formed by values of $C_{out}$ that are multiples of 4.
These configurations are executing between $17-20\%$ faster than the rest of the configurations.

The previous study presents similar results for this setup, with execution time increasing stepwise at approximately the same rate. Compared to our estimates, the original results overestimate the actual execution time of the convolutional configurations.

\begin{figure*}[t]
    \centering
    \begin{subfigure}[b]{0.48\textwidth}
        \centering
        \includegraphics[width=\linewidth]{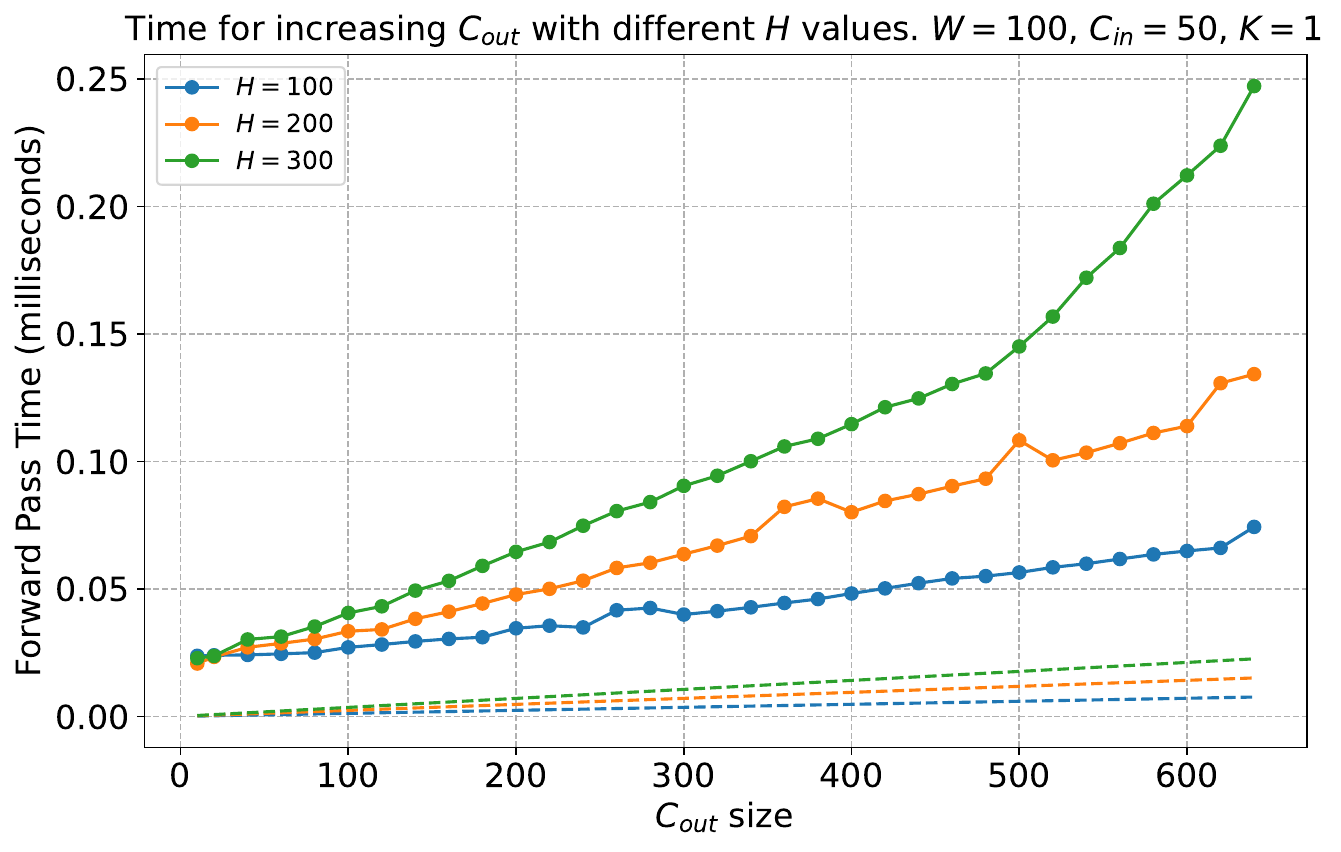}
        \caption{Limited sampling: 33 datapoints}
        \label{fig:exp-F}
    \end{subfigure}
    \hfill
    \begin{subfigure}[b]{0.48\textwidth}
        \centering
        \includegraphics[width=\linewidth]{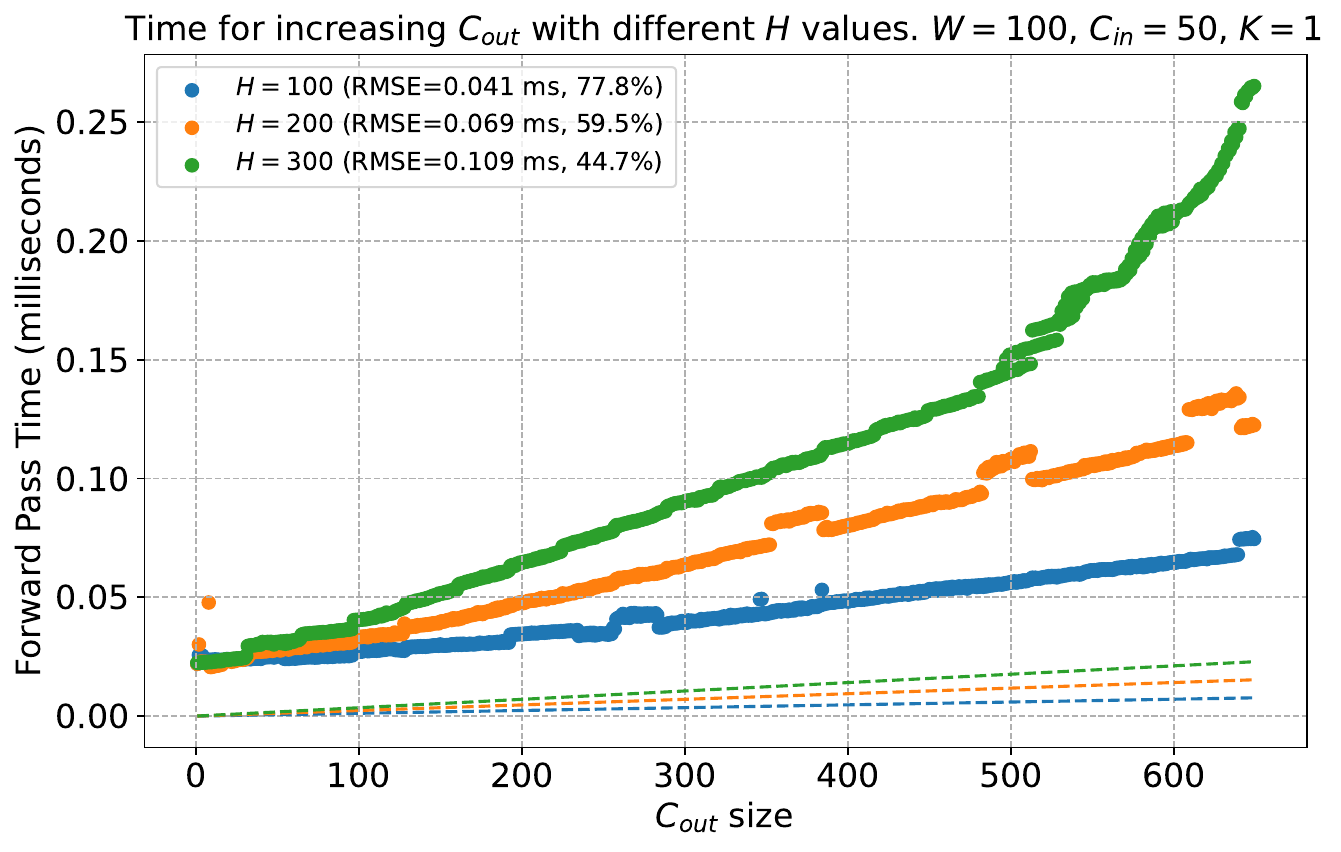}
        \caption{Full sampling: 649 datapoints}
        \label{fig:exp-FE}
    \end{subfigure}
    \caption{Experiment F: Execution time and estimations for increasing output channels and different input sizes. Estimations are represented with the dashed line.}
    \label{fig:exp-F-combined}
\end{figure*}

\autoref{fig:exp-F-combined} displays the results of Experiment F. 
In this experiment, the independent variable is again the number of output channels $C_{out}$, but tested for different input heights $H$, with a kernel size $K=1$.
\autoref{fig:exp-F} presents the results for the limited sampling of 33 datapoints. In this configuration, the forward pass time increases consistently with both the number of output channels $C_{out}$ and the input height $H$. 
The accuracy of the $\alpha-FLOPs$ estimation is much worse than seen in previous experiments. 
Here, the formula underestimates execution time by a wide margin across all three sets of configurations, with values close to $0$. Normalized error ranges from $48\%$ to as high as $82\%$ of the actual measurements

The full-resolution results in \autoref{fig:exp-FE} reveal less instability than those of previous experiments in this category. For $H=300$, the execution time follows a linear trend, but shows a significant jump in latency after $Cout$ exceeds 500. In contrast, for H=100 and H=200, the execution time remains relatively linear across wide ranges of $C_{out}$, with some jumps in latency at certain values that return to the original trend.

These results are slightly more stable than those from the original experiments. 
The original study shows approximately linear behavior, with small step-ups similar to those observed in the previous experiments, though smaller in magnitude.

\begin{figure*}[t]
    \centering
    \begin{subfigure}[b]{0.48\textwidth}
        \centering
        \includegraphics[width=\linewidth]{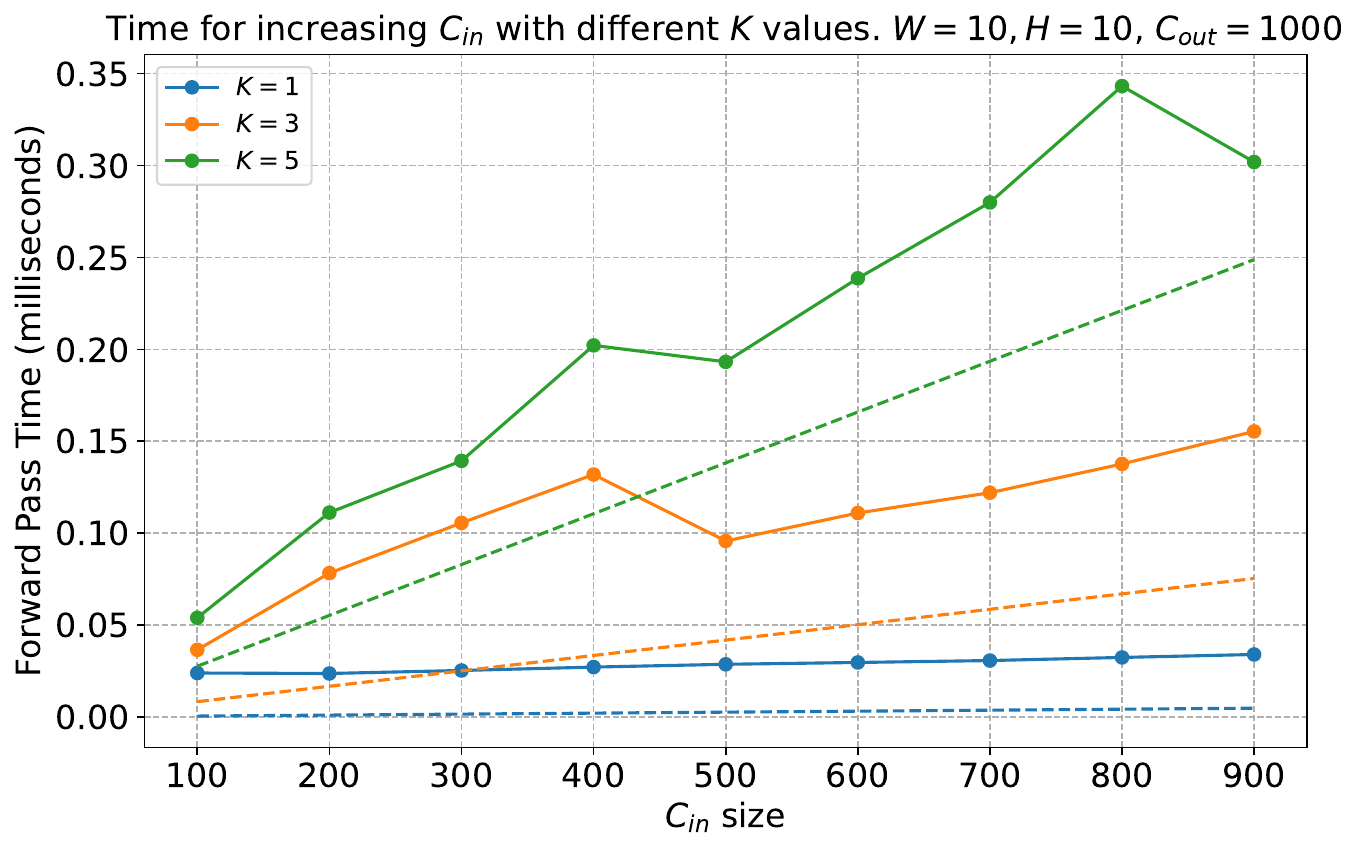}
        \caption{Limited sampling: 9 datapoints}
        \label{fig:exp-G}
    \end{subfigure}
    \hfill
    \begin{subfigure}[b]{0.48\textwidth}
        \centering
        \includegraphics[width=\linewidth]{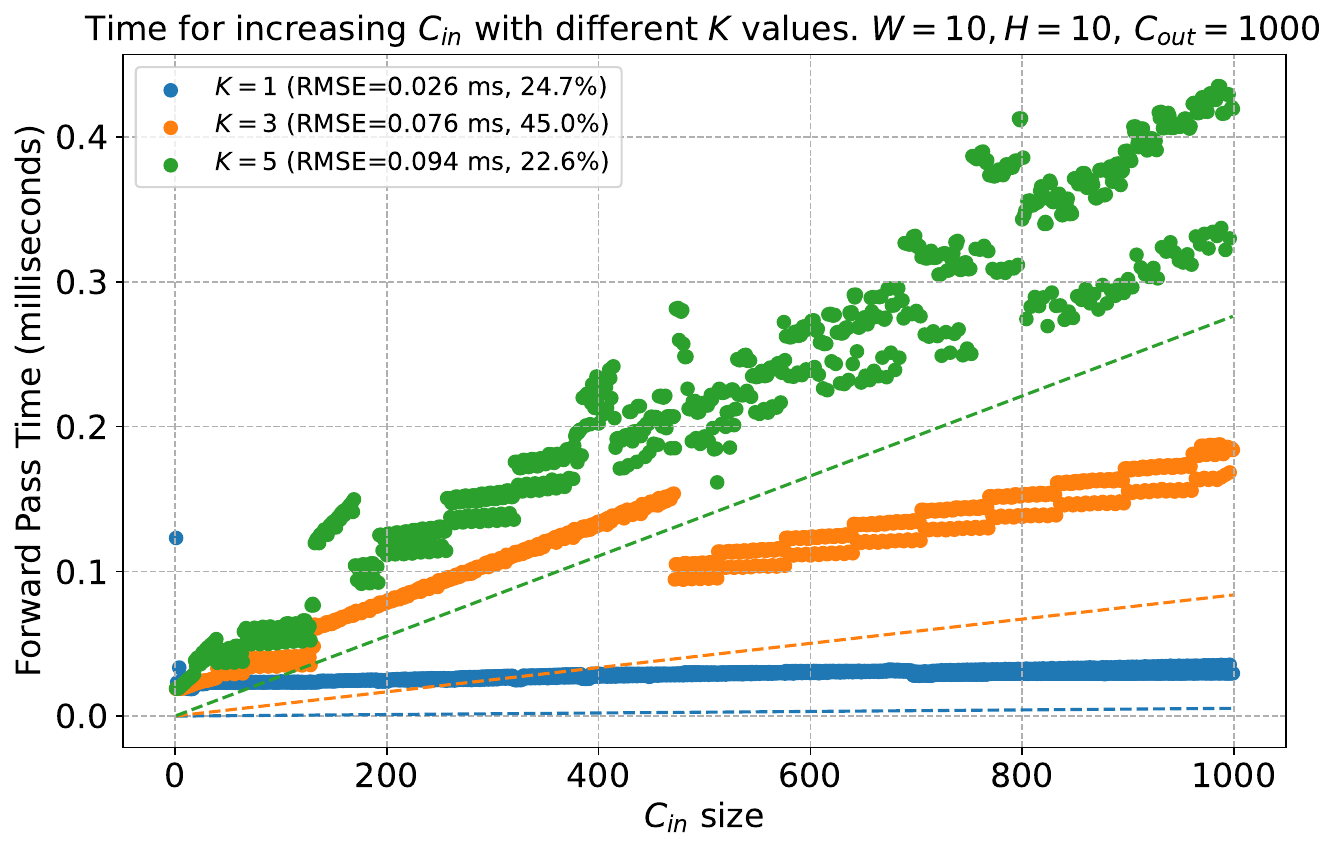}
        \caption{Full sampling: 1000 datapoints}
        \label{fig:exp-GE}
    \end{subfigure}
    \caption{Experiment G: Execution time and estimations for increasing input channels and different kernel sizes. Estimations are represented with the dashed line.}
    \label{fig:exp-G-combined}
\end{figure*}

\autoref{fig:exp-G-combined} displays the results of Experiment G. In this scenario, the independent variable is the number of input channels $C_{in}$ , tested across three different kernel sizes $K \in \{1,3,5\}$ , while maintaining a small spatial resolution of $H=W=10$.  

\autoref{fig:exp-G} illustrates the low-resolution sampling with only 10 datapoints. The execution time for $K=1$ is nearly negligible and remains constant regardless of $C_{in}$ size. For larger kernels, the execution time increases but shows unexpected non-linearities.
For example, execution time suddenly improves at $C_{in}=500$ for both $K=3$ and $K=5$. The estimation formula fails to capture these hardware-specific variations and generally underestimates the computational cost for K=3 and K=5.  

\autoref{fig:exp-GE} presents the full sampling with 1000 datapoints, and reveals vast instabilities that are not visible with the original resolution. 
The instability is highly pronounced for $K=5$. For lower values of $C_{in}$, the execution time of these configurations divides roughly into two bands, with multiples of 4 executing slightly faster, as seen for experiment E.
However, for values above $C_{in} \approx 400$, measurements are widely scattered across multiple efficiency tiers. 
Another discontinuity in execution time is visible for $K=3$ at $C_{in} \approx 480$, where the execution time abruptly drops, and the two previously mentioned efficiency bands become visible.

\section{Discussion}

\subsection{Estimating time through FLOPs}

Based on the results and our replication objectives, we achieved a partially successful replication.
With respect to \textbf{RO1}, we replicate the empirical results showing that, in our setup, FLOPs is not a reliable proxy for convolutional-layer execution time.
The first three experiments (Experiments A, B, and C), in which FLOPs are held constant across configurations, show that execution times for configurations with the same FLOPs are not always equal.
Significant variance in execution time is introduced depending on which dimensions are increased and how effectively the GPU can parallelize these operations.
In general, as observed in our results and in those of the previous paper, spatial dimensions are more easily parallelized than kernel dimensions.
The results from the Increasing FLOPs experiments (Experiments D through G) also show that execution time does not always increase linearly with FLOPs, despite the theoretical equation for FLOPs suggesting the opposite.
The results of our replication experiments validate the thesis of previous work \cite{flopsCriticism} that using raw FLOPs as a direct metric for execution time or the computational cost of a convolutional layer is not appropriate.

Despite these findings being available for several years, FLOPs is still a widely used metric when reporting computational costs and model complexity of AI models, both in Computer Vision with CNNs~\cite{8100126, pmlr-v238-meng24a, DBLP:journals/corr/abs-2103-10858, DBLP:conf/cvpr/SainMCKBS25} and in newer fields such as LLM research~\cite{DBLP:conf/naacl/LiuSHWWZJCHQ22, jiao-etal-2020-tinybert, liu-etal-2020-fastbert, li-etal-2021-accelerating, DBLP:journals/jmlr/FedusZS22, pmlr-v162-du22c, huang2024towards, 10.1145/3677375}.
This happens not only in research, but also for commercial products or platforms for model sharing, such as HuggingFace \cite{different_flops_2026}.
The reason is simple: FLOPs is an easy-to-calculate metric that can be computed even without running the model, and intuitively, the more operations an AI model needs to perform, the more time, energy, and resources it takes.
As shown by these results, practitioners need to be careful when doing these approximations.
    
Considering our \textbf{RO2}, we partially replicate the overall trend from the original study: spatial FLOPs are generally easier to parallelize than kernel FLOPs.
However, the results of the experiments also reveal that the relationship between FLOPs and execution time is much less straightforward than shown in the previous study.
Fine-grained measurements for different values of the independent variables show that efficiency can vary greatly between consecutive values.
Execution time, in some cases, evolves in jumps rather than uniformly and shows patterns that depend on input dimensions.
Many of these jumps or behaviors appear in multiples of certain numbers, such as 4, 32, 41, or 64, as shown in experiments D and E.

The most likely explanation is that the patterns derive from both hardware and software characteristics.
GPUs perform multiple optimizations to improve parallelization.
Some examples are Spatial Tiling for caching \cite{VANWERKHOVEN201414} or thread parallelization with streaming multiprocessors \cite{electronics8121451}.
The GPU's specs, such as cache, memory, or bus sizes, also play a role.
The RTX 4090 has 72 MB of L2 cache and a 384-bit bus \cite{10579250}.
A small increase in input size can cause the cache or bus to fill up, requiring additional memory reads and/or writes.

Software causes for these patterns can be differences in the implementations of convolution algorithms \cite{8966288}.
The NVIDIA CUDA libraries include multiple algorithms for convolution operations.
Some of these implementations prioritize computational efficiency or memory efficiency.
Depending on input size, the CUDA libraries can select one of the multiple implementations, which can affect execution time.

Finally, with respect to \textbf{RO3}, we do not manage to replicate accurate estimations using the $\alpha-FLOPs$ approach.
The $\alpha-FLOPs$ formula is an attempt to mathematically model the non-linearity observed in the convolutional experiments.
The formula accounts for the greater parallelization of spatial dimensions and the additional costs associated with higher kernels.
Our results show limited success in this area.
While the estimations work well for a couple of experiments, most configurations are underestimated.

The limitations of the formula are evident in its components.
The correction only considers the input size $S$ and kernel size $K$, but does not account for the impact of input and output channels.
Our results show that these two factors play an important role in execution time for our setup, exceeding the effect of the single observer in the original study.
If the authors did not observe these effects in their hardware, they did not account for them, making the formula obsolete for newer GPUs.

Potentially, the $\alpha-FLOPs$ formula could be extended and adapted, based on the novel behaviors observed in our results, including variables to account for the effect of input and output channels.
However, it would not be possible to design a formula that accounts for edge cases related to the software and hardware optimizations we have discussed, due to the number of dependencies introduced by different software libraries or hardware configurations.

\subsection{Replication Material}





Ultimately, we partially reproduced the results of the previous study.
Some of the reasons for replication failures are likely due to differences in hardware.
However, the lack of replication materials in the original paper significantly undermines the results.
The multiple assumptions we had to make about some of the authors' choices introduce confounding factors, preventing us from separating which differences arise from hardware or from setup discrepancies.

For the empirical measurements, we obtain values that are consistently $\sim 10\times$ faster than in the original study.
The main reason for this speedup is differences in GPU hardware, including available cores, memory, and compute units.
For example, the Ada Lovelace architecture of the RTX 4090 includes compute units to perform operations on TF32 floating-point numbers, a format with a reduced mantissa designed for AI workloads \cite{10.1007/978-3-031-97554-7_11}, which is not available in the Turing architecture in the original paper.
This format can give up to $3-4\times$ speedup for AI workloads.
Combined with other architectural improvements, the hardware factor is more likely to explain the overall speedup observed in the measurements.

Aside from the time measurement magnitudes, the behavior observed in the experiments for the different convolutional configurations on our hardware only partially resembles the original results.
For Experiments A, E, and F, the trends observed for execution time are similar to the original results.
For the remaining experiments, while trends are mostly similar, we obtain more unstable results.
This might be explained again by differences in hardware.
This is likely the case for experiments B and C, where we used the exact same values as the original study.

However, thanks to the higher resolution of the independent variable in the Increasing FLOPs experiments, we observe high execution-time variance across very similar convolutional configurations.
Using the lower-resolution version from the original paper, this variance is hidden, and the results are much closer to the original study.
Therefore, it is not possible to determine whether this variance was also present in the hardware used in the original paper or was introduced by our newer architecture.

The main negative result in our replication is the usage of $\alpha-FLOPs$ to estimate energy consumption. 
Except for two experiments, the estimates fall short of the actual time values.
There are two possible reasons to explain this.
First, the formula itself might not adapt well to the significantly faster timings observed in our GPU.
The formula is not linear, as it includes polynomial terms due to $\gamma_k$, $K$, and $S$ in the denominator.
Therefore, the formula might lose resolution for lower timings.

A second reason for the negative results in the estimation can be that we did not perform the regression as the original authors did.
In this scenario, the results could potentially be reproduced, but since the replicability package was missing, some of the assumptions we made could be incorrect, tainting the results.
The assumptions are threats to the validity of the replication and could have affected the results in the following way.

\textbf{Assumption 1}: We define the range of hyperparameters for the regression dataset based on the experiments to replicate.
Since the search space for all parameter combinations is very large, we had to select a sampling strategy.
It is possible that the sampling strategy we chose differs too much from the original sampling.
Given the high variance observed during the measurements, it is possible that the distribution of time measurements in the regression dataset differs significantly from that used in the original paper.

\textbf{Assumption 2}: The original study provided vague details regarding the non-linear regression implementation. 
While we set up the configuration as close as possible to the original paper, by using the \textit{scipy.curve\_fit} library and setting specific initial values, our replication for the regression might have converged on a different local minimum. 
This could have contributed to the consistent underestimation of execution times observed in the experiments.

\textbf{Assumption 3}: In the original dataset, a hardcoded $final$ scaling factor was used without justification in the $\alpha-FLOPs$ computation.
Treating this as an additional regression parameter in the replication was intended to allow the formula to adapt to the RTX 4090's significantly higher computing power, rather than reusing values without justification.
However, this additional degree of freedom may have allowed the model to overfit to the average latency of the regression set, masking the formula's inability to capture how individual hyperparameters—specifically kernel size and channel counts—interact on newer GPU architectures.

\textbf{Assumption 4}: Our choice of library versions probably had an effect on time measurements for both the regression dataset and experiments, due to the different optimizations introduced since the writing of the original study.


Ultimately, this paper highlights the need for complete and accurate replication packages for research on efficiency assessment, especially when this efficiency depends heavily on hardware.
A proper replication package would have removed the need for our assumptions, and all of the negative results observed could be directly attributed to hardware differences.
This would largely facilitate the reuse of techniques and results.
For replication packages that yield positive results, techniques like this could be easily reused and extended, and for negative results, it would be easier to identify outdated components due to hardware evolution and propose potential fixes.

For these reasons, we provide a complete replication package for our implementation of the $\alpha-FLOPs$ methodology.
The original approach has the potential to provide reliable estimations.
Our intention is to facilitate other researchers' study of these empirical results and the estimation approach on their own hardware, so extensions and fixes for different kinds of hardware can be developed.

\section{Related Work}
\textbf{FLOPs as proxy for AI model efficiency}.
The validity of floating-point operations (FLOPs) as a proxy for AI model training time and energy consumption remains an ongoing debate in the field.
FLOPs were first proposed as an energy proxy in the early stages of Green AI \cite{10.1145/3381831, DBLP:journals/corr/abs-1910-09700} because they can be easily calculated solely from architectural details.
\citeauthor{flopsToEnergy}~\cite{flopsToEnergy} realizes a study on inference time and energy consumption for a set of AI models for Computer Vision and Natural Language Processing tasks.
There, the authors propose using FLOPs as a proxy for computational efficiency, relating the FLOPs a model uses during inference to hardware FLOPs and combining them with GPU hardware specifications, specifically TFLOP/s and power.
However, their study fails to address the gap between logical FLOPs and hardware FLOPs, as it does not account for potential hardware optimizations introduced by GPUs or TPUs.

Besides the work of \citeauthor{flopsCriticism} that is the main focus of this replication, other works also address this mismatch.
The work of \citeauthor{Chen_2023_CVPR}~\cite{Chen_2023_CVPR} reports that reducing FLOPs does not necessarily yield a similar reduction in latency and proposes a novel family of neural networks that, instead of minimizing FLOPs, optimizes their parallelization to extract spatial features more efficiently.
The proposed neural networks achieve $36\%$ higher inference throughput on GPU, and are up to $2.8 \times$ faster than comparable neural networks.
Compared to our work, this paper does not focus on execution-time estimation or reporting, but rather on exploiting higher parallelization.

\citeauthor{10.5555/3455716.3455964} \cite{10.5555/3455716.3455964} also investigates this issue for the energy consumption of image classification algorithms. 
In their paper, they repeatedly run inference on pre-trained image classification models and measure FLOPs and energy usage, finding little correlation between the two when comparing different architectures but a significant correlation when comparing models within the same architecture.
They determine that FLOPs can be useful to compare relative ordering between architectural classes, but not as a full proxy for energy consumption.
This work relies on energy measurements using a profiler, which is cumbersome and requires access to bare-metal hardware, and again does not propose estimation techniques that are more accessible.

\textbf{Replication of hardware-bound benchmarks}.
One of the main problems with unreliable replication packages is that, in cases of negative reproducibility, it is hard to determine their origin
The work by \citeauthor{10.1145/3736731.3746150} \cite{10.1145/3736731.3746150} describes this problem in HPC benchmarking, where poor replication scripts make it hard to track whether negative results stem from the replication software or from differences in hardware.
They propose an open-source framework to address this challenge, providing a step-by-step methodology for creating replication packages for HPC benchmarks that are repeatable, replicable, and reproducible.
The same principles apply to AI research.
To properly compare the efficiency of AI models across different GPU architectures, replication packages need to be complete and highly reliable, so as to discard any software factors and focus only on hardware differences.

\section{Conclusion}
Our replication study confirms that FLOPs are an inadequate metric for directly assessing the execution time or computational cost of AI models, as efficiency varies significantly depending on how effectively operations can be parallelized across different dimensions. While our experiments on modern hardware (NVIDIA RTX 4090) showed a consistent $10 \times$ speedup compared to the original study’s results, the underlying efficiency trends remained highly hardware-dependent and subject to non-linear "jumps" caused by specific GPU architectures and software optimizations.

With respect to our replication objectives, we provide the following answers:
\begin{itemize}
    \item \textbf{RO1:} We replicate previous findings \cite{flopsCriticism} that FLOPs are not a good proxy for CNN execution time. Across the experiments, configurations with equal FLOPs do not exhibit equal execution time.
    \item \textbf{RO2:} The overall trend from the original study is replicated: spatial FLOPs are generally easier to parallelize than kernel FLOPs. However, on newer hardware we also observe discontinuities and oscillations that were not clearly visible in the original results.
    \item \textbf{RO3:} In our setup, we do not replicate the result of accurate time estimations. The $\alpha$-FLOPs model underestimates latency in most configurations and does not consistently provide accurate estimates.
\end{itemize}

The $\alpha-FLOPs$ estimation methodology showed limited success on newer hardware, consistently underestimating latency across most configurations. 
This failure seems to stem largely from the formula's inability to account for the impact of input and output channels, which played a much more significant role in our setup than in the original study. 
Furthermore, our high-resolution measurements revealed that execution time is far more unstable than previously reported.
However, the lack of original execution code and specific dependency details (such as library and driver versions) forced us to make several assumptions that introduced confounding factors into our replication.
This means that our conclusions about the methodology for newer hardware might be compromised by one of our assumptions. 
Future work on this topic will focus on studying GPU optimization behavior more closely and applying possible fixes to the original approach.

Ultimately, this work validates the empirical findings from the original study but shows negative results when applying the $\alpha-FLOPs$ estimation approach. Our study also highlights the critical necessity for complete and accurate replication packages in AI efficiency research. To foster transparency and enable the community to extend this estimation approach to newer architectures, we provide a fully documented replication package. These findings serve as a call to action for researchers to adopt more rigorous replication standards for efficiency assessment of AI.

\printbibliography

\end{document}

%% file: header.tex
\pdfoutput=1
\usepackage[utf8]{inputenc}

\usepackage[T1]{fontenc}
\usepackage{url}
\usepackage{flushend}
\usepackage{xspace}
\usepackage[skip=5pt]{caption}
\usepackage[inline]{enumitem}
\usepackage{ifthen}
\usepackage{balance}
\usepackage{hyperref}
\usepackage{flushend}
\usepackage{colortbl}
\usepackage{listings}

\definecolor{codegreen}{rgb}{0,0.6,0}
\definecolor{codegray}{rgb}{0.5,0.5,0.5}
\definecolor{codepurple}{rgb}{0.58,0,0.82}
\definecolor{backcolour}{rgb}{0.95,0.95,0.92}

\lstdefinestyle{mystyle}{
    commentstyle=\color{codegreen},
    keywordstyle=\color{magenta},
    numberstyle=\tiny\color{codegray},
    stringstyle=\color{codepurple},
    basicstyle=\ttfamily\footnotesize,
    breakatwhitespace=false,         
    breaklines=true,                 
    captionpos=b,                    
    keepspaces=true,                 
    numbers=left,                    
    numbersep=5pt,                  
    showspaces=false,                
    showstringspaces=false,
    showtabs=false,                  
    tabsize=2
}

\usepackage{microtype}
\usepackage{multirow}
\usepackage[]{siunitx}
\newcommand*\np[2][z]{
\ifx z#1%
$\num{#2}$%
\else%
$#2\ #1$
\fi\xspace%
}
\usepackage{fp}
\usepackage{xfp}

\usepackage{graphicx}
\usepackage{subcaption}
\usepackage{pifont}

\usepackage{booktabs}
\usepackage{multirow}
\usepackage{threeparttable}
\usepackage{xcolor}
\usepackage{array,makecell}
\usepackage{diagbox}
\usepackage{tabularx}

\usepackage{mathtools}
\usepackage{amsmath}

\usepackage{tcolorbox}
\usepackage{pgfplots}
\usepackage{pgfplotstable}

\usepackage{enumitem}
\usepackage[framemethod=TikZ]{mdframed}
\usepackage[scaled]{beramono}

\usepgfplotslibrary{statistics}
\usetikzlibrary{calc,patterns,fpu} 
\pgfplotsset{compat=1.14} 

\pgfplotsset{
    /pgf/declare function={
        Floor(\x) = round(\x-0.49);
    },
    show sum on top/.style={
        /pgfplots/scatter/@post marker code/.append code={%
            \path let \p1=($(normalized axis cs:%
                        \pgfkeysvalueof{/data point/x},%
                        \pgfkeysvalueof{/data point/y})%
                        -(normalized axis cs:\pgfkeysvalueof{/data point/x},0)$)
            in node[
                at={(normalized axis cs:%
                        \pgfkeysvalueof{/data point/x},%
                        \pgfkeysvalueof{/data point/y})%
                },
                anchor={-90*sign(\y1)},yshift={sign(\y1)*2pt}
            ]
            {\pgfmathprintnumber[fixed, precision=1]{\pgfkeysvalueof{/data point/y}}};
        },
    }
}

\definecolor{blau_1a}{RGB}{93,133,195}
\definecolor{blau_2a}{RGB}{0,156,218}
\definecolor{gruen_3a}{RGB}{80,182,149}
\definecolor{gruen_4a}{RGB}{175,204,80}
\definecolor{gruen_5a}{RGB}{221,223,72}
\definecolor{orange_6a}{RGB}{255,224,92}
\definecolor{orange_7a}{RGB}{248,186,60}
\definecolor{rot_8a}{RGB}{238,122,52}
\definecolor{rot_9a}{RGB}{233,80,62}
\definecolor{lila_10a}{RGB}{201,48,142}
\definecolor{lila_11a}{RGB}{128,69,151}

\definecolor{blau_1b}{RGB}{0,90,169}
\definecolor{blau_2b}{RGB}{0,131,204}
\definecolor{gruen_3b}{RGB}{0,157,129}
\definecolor{gruen_4b}{RGB}{153,192,0}
\definecolor{gruen_5b}{RGB}{201,212,0}
\definecolor{orange_6b}{RGB}{253,202,0}
\definecolor{orange_7b}{RGB}{245,163,0}
\definecolor{rot_8b}{RGB}{236,101,0}
\definecolor{rot_9b}{RGB}{230,0,26}
\definecolor{lila_10b}{RGB}{166,0,132}
\definecolor{lila_11b}{RGB}{114,16,133}

\newboolean{showcomments}
\setboolean{showcomments}{true}

\input{ChartInTable.tex}

\definecolor{mygray}{RGB}{240,240,240}

\ifthenelse{\boolean{showcomments}}
 { }
        { }

\definecolor{eminence}{RGB}{108,48,130}
\definecolor{weborange}{RGB}{255,165,0}
\definecolor{frenchplum}{RGB}{129,20,82}
\definecolor{darkgreen}{RGB}{10, 92, 10}

\mdfdefinestyle{mpdframe}{
    nobreak                     =true,
    backgroundcolor             =black!3,
    frametitlebackgroundcolor   =black!15,
    frametitlerule              =false,
    roundcorner                 =5pt,
    innermargin                 =0.3cm,
    innerleftmargin             =0.3cm,
    innerrightmargin            =0.3cm,
    innertopmargin              =0.3cm,
    innerbottommargin           =0.3cm,
    shadowsize                  =1pt
}

%% file: ChartInTable.tex
\newcommand{\ShowAbsoluteNumber}[1]{%
\ifnum #1<10%
{\hspace*{0pt}#1}%
\else%
\ifnum #1<100%
{\hspace*{0pt}#1}%
\else%
\ifnum #1<1000%
{\hspace*{0pt}#1}%
\else%
{\numprint{#1}}%
\fi%
\fi%
\fi%
}

\newcommand{\ShowPercentage}[2]{%
\FPeval\percentage{round(#1/#2*100,0)}%
\FPeval\percentageOneDecimal{round(#1/#2*100,1)}%
\ifnum \percentage=0%
{\np[\%]{\FPprint{percentageOneDecimal}}}%
\else%
\ifnum \percentage<10%
{\np[\%]{\FPprint{percentageOneDecimal}}}%
\else%
{\np[\%]{\FPprint{percentageOneDecimal}}}%
\fi%
\fi%
\xspace
}
\newlength\BARSIZE  
\newcommand{\inlinechart}[2]{%
\FPeval{\BLACKBARSIZE}{#1/#2}\textcolor{black!80}{\rule{\BLACKBARSIZE\BARSIZE}{1.6ex}}%
\FPeval{\BLACKBARSIZE}{1 - (#1/#2)}\textcolor{black!10}{\rule{\BLACKBARSIZE\BARSIZE}{1.6ex}}%
}

\newcommand*\percent[3][v]{%
\ifx q#1%
    \np{#2}/\np{#3}(\ShowPercentage{#2}{#3})\else%
\ifx s#1%
    \ShowPercentage{#2}{#3}\else%
\ifx p#1%
    \np{#2}(\ShowPercentage{#2}{#3})\else%
\ifx c#1%
    \inlinechart{#2}{#3}%
\else%
    \np{#2}%
    \ifx r#1%
        /\np{#3}%
    \fi%
    \hspace*{0.5ex}(\ShowPercentage{#2}{#3}) %
    \inlinechart{#2}{#3}%
    \xspace
\fi\fi\fi\fi%
}

%% file: methodology.tex
\section{Methodology}
Our replication is based on the following three steps defined in the original study: (1) data collection of execution times for different convolutional configurations; (2) regression over the collected data to determine the parameters for the $\alpha-FLOPs$ formula in our hardware; (3) time measurements for the experiments reported in the original paper, with estimations from the $\alpha-FLOPs$ formula.

This section introduces the parameter settings for the three steps and the experimental setup.
These settings follow the instructions in the paper and the provided dataset as closely as possible.
When details from available sources are lacking, we make assumptions when selecting parameters and report them in the text.

\subsection{Parameter setting}

\paragraph{Data collection}
Using our own GPU, we collect execution time data for a set of convolutional configurations based on the following sampling strategy:
\begin{itemize}
    \item Input and output sizes $W, H \in \{1, 2, 4, 8, 16, 32, 64, 96, 128, 256\}$. We test every possible combination, not only square matrices. Input and output have the same dimensionality, as per the original paper.
    \item Input and output channels $C_{in}, C_{out} \in\{10 \cdot x\}, x \in \left[1,9\right] \cup \{100 + 50 \cdot x\}, x \in \left[0,20\right] \cup \{2^x \cdot 100\},x \in \{4,5\}$. We select these values because they fall within the range of channel sizes tested across the different experiments.
    \item Kernel sizes $K \in \left[1, 14\right]$. We test only for square kernels, since all the original experiments did so, although this is not explicitly defined in the original paper.
\end{itemize}
To obtain execution time, we run each configuration 50 times and take the average execution time.

\textbf{Assumption 1:} The minimum and maximum values selected for the regression were selected based on the experiments shown in the original paper.
These ranges cover most of the configurations that will be later analyzed during the experiments.
The replication material does not provide any information on the measurement ranges used for their regression, neither in the paper nor in the dataset. 

\paragraph{Configuration for the regression}
To perform the regression using the $\alpha-FLOPs$ formula, we use the least-squares approach.
This approach minimizes the Mean Squared Error between measurements and an arbitrary function, in this case, the $\alpha-FLOPs$ formula.
We define the parameters to fit $\beta_k$ and $\gamma_k$. We also add an additional scaling factor, $final_k$, as a regression parameter, which multiplies the original formula.
The updated formula used for regression looks like
\begin{equation*}
    \alpha_k(S) = final_k \cdot \left(\frac{S_k + \beta_k(S-S_k)}{S}\right)^{\gamma_k}
\end{equation*}
To perform the regression, the function used is \textit{curve\_fit} from the \textit{scipy} package. 
It requires initial parameter values, as well as lower and upper bounds.
Based on the values from the original paper, we set up the initial values to $\beta_k = 0.01$, $\gamma_k = 0.8$, and $final_k = 0.3$. 
We define lower and upper bounds for all three parameters to $[0,2]$.
Similarly to the original paper, we perform two separate regressions: one for convolutional layers with $K=1$ and another for $K>1$.

\textbf{Assumption 2}: We choose the non-linear least-squares based on the regression definition given in the paper and email communication with the authors.
We believe this approach is the one used for the study, but it remains vague in the original material.
The \textit{curve\_fit} function from the \textit{scipy} package was chosen based on data-science standards, and the initial values and bounds were chosen based on the parameter values from the original paper.

\textbf{Assumption 3}: The $final_k$ parameter is added as a mirror to the one hardcoded in the provided dataset. Since the original value does not appear in the paper and is not justified in the original code, it is more appropriate to treat it as an additional regression parameter rather than to reuse or manually adjust it.



\input{table}

\paragraph{Measurement set}
Table \ref{tab:experiments} shows the different experiments run in this replication. Each experiment measures execution time for a forward pass across different convolutional configurations, following specific rules for the hyperparameters. 
The experiments can be divided in 2 groups:
\begin{itemize}
    \item \textbf{Constant FLOPs}: For these experiments, the variables are configured such that some hyperparameters are increased while others are decreased, following a proportion that keeps the FLOPs constant.
    This effectively changes the parallelization capabilities between different configurations.
    The objective of these experiments is to determine whether configurations with the same FLOPs have the same execution time, regardless of parallelization.
    These are experiments A, B, and C.
    \item \textbf{Increasing FLOPs}: In these experiments, two hyperparameters are increased, while the rest are kept constant.
    These experiments examine the shape of execution time as certain dimensions are increased, to determine how each dimensions affect final execution times
    These are experiments D, E, F and G.
\end{itemize}

For each experiment, we run each configuration 2000 times and report the average execution time to remove noise.
The configurations for each experiment are taken directly from the dataset provided with the original manuscript, but we extend some of them to make use of the increased capabilities of our hardware.

Experiment A is extended to test one additional power of 2 in the input channels compared to the original study.

For the Increasing FLOPs experiments, we measure the whole range of the variable. This is, from 1 to 650 in D, E, F, and from 1 to 1000 in G.
In the original study, the authors limited their testing to a sample of this space, with equidistant data points, 32 for D, E, F, and 10 for G.

\subsection{Experimental setup}
\paragraph{Hardware}
Our replication is performed on an AMD Ryzen 7900X CPU and an Nvidia RTX 4090 GPU.
The 4090 was launched in October 2022.
It has a maximum clock speed of 2520 MHz, with 16,384 CUDA cores. It also has a larger L2 cache of 72MB and 24GB of GDDR6X VRAM.

Comparatively, the original experiments are run on an Intel Core i7-9850H CPU and a NVIDIA Quadro T2000.
The Quadro T2000 is a mobile GPU for laptops and is older than the 4090, launched in May 2019. It has 1,024 CUDA cores, with a maximum frequency of 1785 MHz.
It also has 1MB of L2 cache and 4GB of GDDR5 VRAM.


\paragraph{Library versions}
For our replication, we choose PyTorch version 2.9, with CUDA library 12.2.
The choice is based on PyTorch's current popularity and the currently available versions for it and CUDA.
This provides a perspective on CNN efficiency in the current state of the art, both hardware- and software-wise, and closer to a real use scenario.

\textbf{Assumption 4}: We could not identify the Deep Learning library used for the experiments, nor the version of the CUDA libraries, from the manuscript or the dataset.
These versions are likely far from the original versions in the paper.
Since there is no reference in the original library, we choose library versions that give a view of computational efficiency in current-day scenarios.

%% file: table.tex
\begin{table*}[h]
\begin{tabular}{lllllll}
\bottomrule
Experiment                                       & $H$               & $W$                  & $K$                       & $C_{in}$                           & $C_{out}$  &  Original Figure \cite{flopsCriticism}                                  \\ \bottomrule
\multicolumn{1}{l}{\multirow{3}{*}{A}} & $1$               & $2$                  & $1$                       & $\mathbf{2^x \cdot 100, x \in \left[0,9\right]}$                    & $\mathbf{2^x \cdot 50, x \in \left[0,9\right]}$  & \multirow{3}{*}{Figure 3b}\\
\multicolumn{1}{c}{}                           & $2$               & $2$                  & $1$                       & $\mathbf{2^x \cdot 50, x \in \left[0,9\right]}$                     & $\mathbf{2^x \cdot 50, x \in \left[0,9\right]}$ \\
\multicolumn{1}{c}{}                           & $4$               & $4$                  & $1$                       & $\mathbf{2^x \cdot 25, x \in \left[0,9\right]}$                     & $\mathbf{2^x \cdot 25, x \in \left[0,9\right]}$ \\ \bottomrule
B                                        & $\operatorname{round}(300/K)$           & $\operatorname{round}(300/K)$              & $\mathbf{\left[1,30\right]}$ & $\{150,100,70,50\}$                & $C_{out} \leftarrow C_{in}$ & Figure 4a \\ \bottomrule
C                                       & $10$              & $10$                 & $\mathbf{\left[1,30\right]}$       & \makecell[l]{$\operatorname{round}(R/K),$ \\ $R \in \{4500,3000,2100,1500\}$} & $C_{out} \leftarrow C_{in}$ & Figure 4b           \\ \bottomrule
D                                       & $100$             & $\mathbf{\left[1,650\right]}$ & $3$                       & $\{50,100,150\}$                   & $100$ & Figure 5a                   \\ \bottomrule
E                                        & $100$             & $100$                & $3$                       & $\{50,100,150\}$                   & $\mathbf{\left[1,650\right]}$ & Figure 5b                          \\ \bottomrule
F                                        & $\{100,200,300\}$ & $100$                & $1$                       & $50$                               & $\mathbf{\left[1,650\right]}$ & Figure 6a                           \\ \bottomrule
G                                       & $10$              & $100$                & $\{1,3,5\}$               & $\mathbf{\left[1,1000\right]}$               & $100$ & Figure 6b                                           \\
\bottomrule
\end{tabular}
\caption{Parameter setup for the different experiments. The independent variable is marked in bold. We also indicate to which figure in the original paper each experiment corresponds.}
\label{tab:experiments}
\end{table*}